\documentclass[runningheads]{llncs}
\usepackage{graphicx}
\usepackage{amsmath}
\usepackage{algorithm, algorithmic}

\begin{document}
%
\title{CORPS: Cost-free Rigorous Pseudo-labeling based on Similarity-ranking for Brain MRI Segmentation}
\titlerunning{CORPS for Brain MRI Segmentation}
%
\author{Can Taylan Sari \and
Sila Kurugol \and
Onur Afacan \and
Simon K. Warfield}
\authorrunning{Can Taylan Sari et al.}
%
\institute{Computational Radiology Laboratory (CRL), Boston Children's Hospital, and Harvard Medical School, Boston, MA, USA \\
\email{\{can.sari\}@childrens.harvard.edu}}
\maketitle              
\begin{abstract}
Segmentation of brain magnetic resonance images (MRI) is crucial for the analysis of the human brain and diagnosis of various brain disorders. The drawbacks of time-consuming and error-prone manual delineation procedures are aimed to be alleviated by atlas-based and supervised machine learning methods where the former methods are computationally intense and the latter methods lack a sufficiently large number of labeled data. With this motivation, we propose CORPS, a semi-supervised segmentation framework built upon a novel atlas-based pseudo-labeling method and a 3D deep convolutional neural network (DCNN) for 3D brain MRI segmentation. In this work, we propose to generate expert-level pseudo-labels for unlabeled set of images in an order based on a local intensity-based similarity score to existing labeled set of images and using a novel atlas-based label fusion method. Then, we propose to train a 3D DCNN on the combination of expert and pseudo labeled images for binary segmentation of each anatomical structure. The binary segmentation approach is proposed to avoid the poor performance of multi-class segmentation methods on limited and imbalanced data. This also allows to employ a lightweight and efficient 3D DCNN in terms of the number of filters and reserve memory resources for training the binary networks on full-scale and full-resolution 3D MRI volumes instead of 2D/3D patches or 2D slices. Thus, the proposed framework can encapsulate the spatial contiguity in each dimension and enhance context-awareness. The experimental results demonstrate the superiority of the proposed framework over the baseline method both qualitatively and quantitatively without additional labeling cost for manual labeling.

\end{abstract}
\section{Introduction}

Quantitative brain magnetic resonance imaging (MRI) analysis, which is a non-invasive way of revealing and evaluating anatomical structures in the human brain, is crucial for understanding the human brain and investigating various neurological disorders. Accurate delineation of the anatomical structures is essential to obtain particular measurements including cortical thickness, volumetrics, and spatial formations of the structures, and exploit these measurements for a precise brain MRI analysis. Manual delineation is still the gold standard but it is both time-consuming and prone to errors as it requires long and intensive examination hours of expert radiologists which is impractical.

Various automated approaches including region growing~\cite{ref_region}, deformable models~\cite{ref_deformable}, single-atlas~\cite{ref_atlas_based,ref_atlas_single} and multi-atlas~\cite{ref_atlas_multi} segmentation methods have been proposed in literature. Recently, the advances in deep learning research have shown great promise as an alternative to segment medical images~\cite{ref_unet,ref_vnet}. The majority of the deep learning approaches employ either patch-based~\cite{ref_efficient3d,ref_tiling} or 2D~\cite{ref_quicknat} deep convolutional neural networks (DCNNs) to segment brain MRI patches or slices, respectively. The aim of training DCNNs on patches/slices is to avoid the computation and memory limitations of the hardware rather than to employ the most accurate model. The performance of these methods may be adversely affected by the deficiencies of spatial contiguity and/or inter-patch adjacencies.



The inadequacy of labeled data and cost of obtaining it for image segmentation led researchers to design methods that utilize unlabeled data to improve the performance of image segmentation tasks. Semi-supervised learning (SSL) methods aim to exploit a large amount of unlabeled data to improve the performance of supervised learning methods which are trained on limited and costly labeled data~\cite{ref_ssl}. These methods can be broadly categorized into consistency-regularizaton~\cite{ref_mean_teacher,ref_temporal,ref_popcorn}, pseudo-labeling~\cite{ref_quicknat,ref_pseudo,ref_weakly}, adversarial~\cite{ref_gan1,ref_gan2}, and active learning~\cite{ref_active} for computer vision and medical image segmentation tasks. 



In this work, we propose CORPS, a semi-supervised segmentation framework built upon a novel atlas-based pseudo-labeling method utilized to train a 3D DCNN for brain MRI segmentation. The contributions of the proposed framework are as follows:

\begin{enumerate}
  \item Inspired by curriculum learning~\cite{ref_curriculum}, which trains machine learning methods on samples in increasing difficulty order, we propose a novel atlas-based pseudo-labeling method that quantifies pairwise similarities between unlabeled and labeled samples and generate pseudo-labels for the unlabeled samples starting from the ones that are most similar to the labeled data. The goal of this labeling strategy is to generate more accurate pseudo-labeled data to include in training to enhance the performance.
  \item Instead of training a multi-class segmentation network, which performs poorly on limited and imbalanced data~\cite{ref_multivsbinary}, we propose to train a separate binary 3D DCNN for each anatomical structure. We employ a lightweight and efficient architecture in terms of the number of filters for binary segmentation of each anatomical structure and do not need extensive computational resources required by highly-parameterized multi-class segmentation networks.
  \item We propose to employ 3D DCNNs trained on full-scale and full-resolution 3D MRI volumes instead of volumes split into 2D/3D patches or 2D slices since the best performance can be achieved using the largest patch size which is the entire volume~\cite{ref_largepatch}. The reduction in the number of filters by utilizing binary segmentation also allows to increase the input size and dimensions. Thus, CORPS can encapsulate the spatial contiguity in each dimension and avoid losing inter-patch adjacency by employing a more context-aware model.
 
  \item We employ the proposed atlas-based method to pseudo-label the unlabeled images in training time and the proposed 3D DCNN to predict the segmentation labels for unseen images in test time. Hereby, our method is not exposed to a long computational time required by the classical atlas-based segmentation methods in test time and is able to provide faster predictions for unseen test samples. This allows not only to exploit the accuracy of atlas-based methods but also the prediction swiftness of deep neural network methods. 
  
  
\end{enumerate}

\begin{figure}
\includegraphics[width=\textwidth]{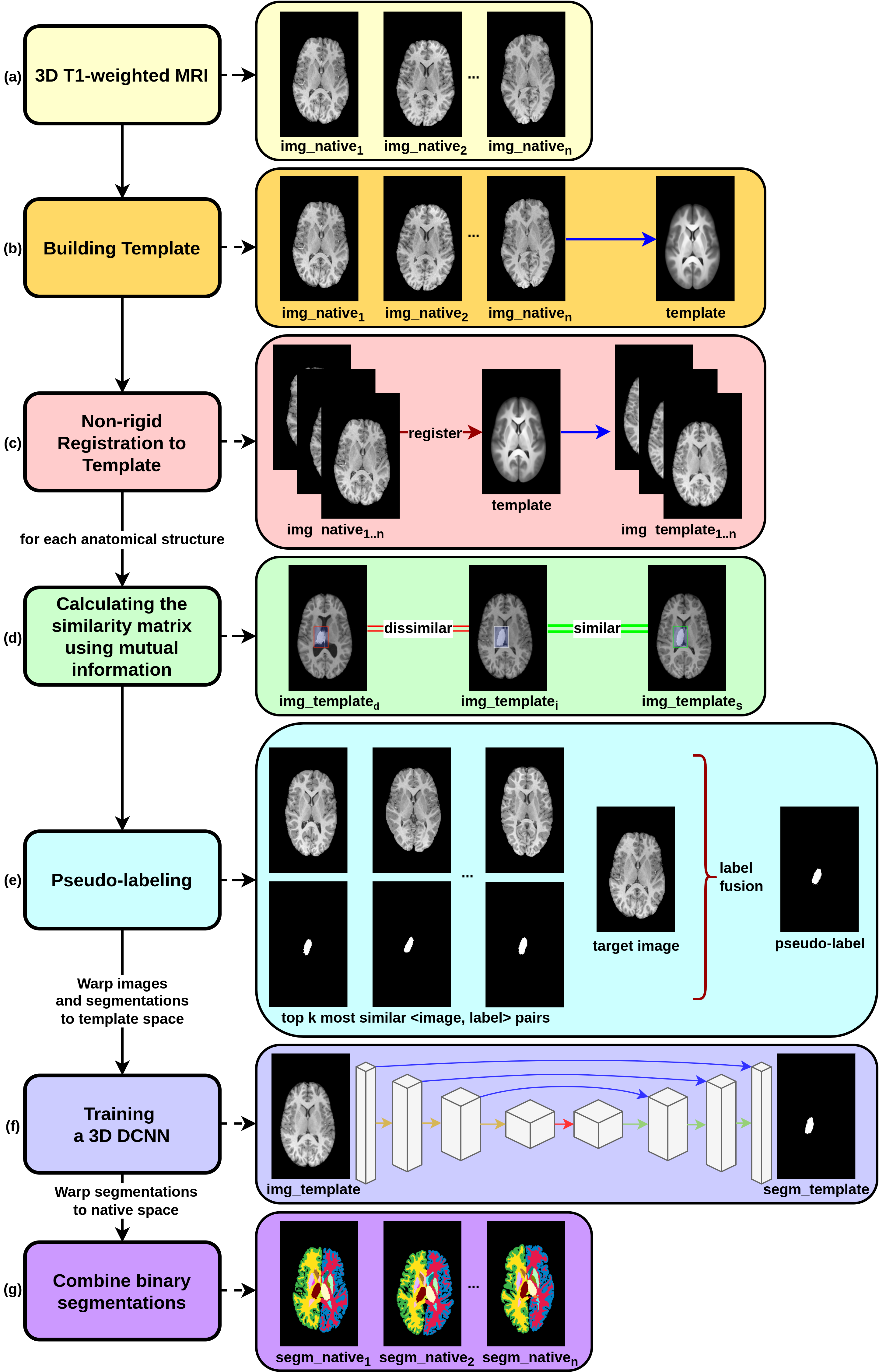}
\caption{Schematic overview of the proposed framework. (d) The expert segmentation label map is overlaid only to improve readability and to emphasize the accuracy of the proposed similarity measure since the method only uses voxel intensity values to calculate the similarities. } \label{fig1}
\end{figure}

\vspace{-0.1cm}
\section{Method}

Our proposed approach relies on improving the performance of the supervised DCNN-based segmentation without additional cost of manual labeling by generating pseudo-labels for unlabeled pool of images using a novel atlas-based method (Fig.~\ref{fig1}). This method proposes to label each unlabeled image using an order based on similarity ranking. To this end, the proposed method first calculates similarities between unlabeled and labeled images for each anatomical structure (Sec.~\ref{sec:similarity}). Afterwards, starting from the most similar unlabeled image to the labeled image pool, each unlabeled image is pseudo-labeled by fusing the expert segmentation labels of the most similar labeled images (Sec.~\ref{sec:atlas}). Finally, a set of 3D DCNNs is trained on the combination of expert and pseudo labeled images to obtain a model that is capable of predicting accurate segmentation labels for each unlabeled test image (Sec.~\ref{sec:dcnn}).

\subsection{Similarity Matrix Calculation}
\label{sec:similarity}
Our proposed pseudo-labeling method hypothesizes that anatomical structures with similar spatial shapes and intensity values should have similar segmentation maps. From this point of view, we define a similarity measure between unlabeled and labeled images for each anatomical structure. We compute similarity score between two images based on the joint mutual information between these two images for a specific anatomical structure. 

The proposed method starts with building a template using unlabeled and labeled images to register images to a common template space. The reason for this registration is to align the anatomical structures to a proximate spatial region and to make it easier and more accurate to calculate the similarities between the images for each anatomical structure. The template is generated with a non-rigid iterative averaging method provided in~\cite{ref_ants} using all images and these images and corresponding segmentation label maps are warped to the template space with a non-rigid registration. Then, we define a bounding box around each anatomical structure in template space since each anatomical structure is substantially aligned in a proximate region. The bounding box for each anatomical structure is obtained by detecting the minimum rectangular volume that encloses all voxels for the anatomical structure in expert segmentation labels of training images. Finally, for each anatomical structure $AS$, the similarity between each unlabeled, $U_i$, and labeled, $L_j$, image is determined by calculating joint mutual information between the bounding boxes, $BU_i$, $BL_j$, defined around the same location in two images as follows:

\begin{equation}
JMI_{AS}(U_i; L_j) = \sum_{x \in BU_i} \sum_{y \in BL_j} p_{(X,Y)}(x,y) log\left(\dfrac{p_{(X,Y)}(x,y)}{p_{(X)}(x)p_{(Y)}(y)}\right)
\end{equation}

The similarity is proportional to the joint mutual information between voxel intensity values of the bounding boxes defined around the structure in each image. Fig.~\ref{fig1} (d) illustrates the level of similarity qualitatively between similar and dissimilar images regarding the left thalamus structure. 


\subsection{Atlas-based Pseudo-labeling for Unlabeled Images}
\label{sec:atlas}

\begin{figure}[t]
\includegraphics[width=\textwidth]{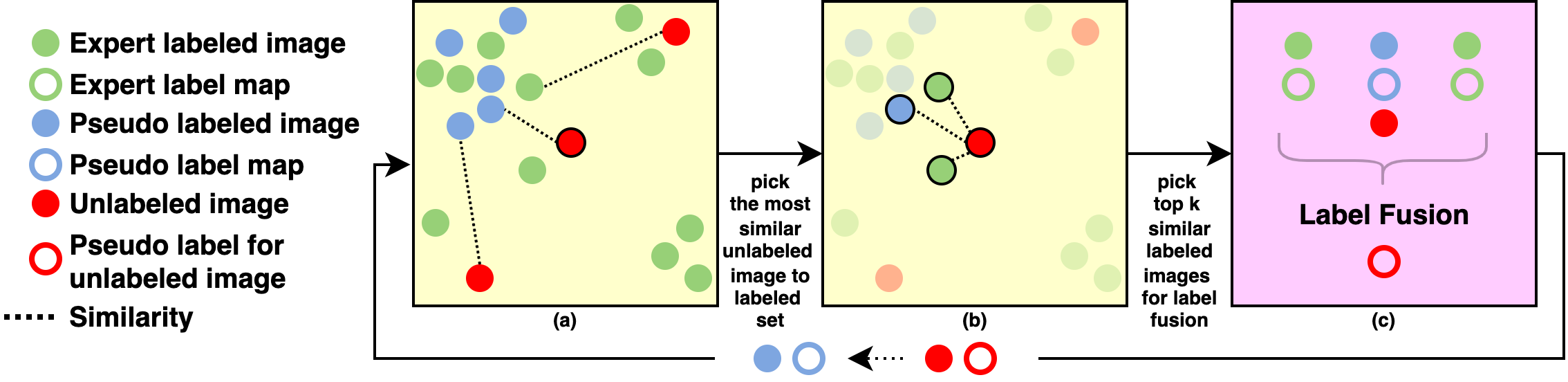}
\caption{(a) The unlabeled image with the highest ${\cal K}$\textsuperscript{th} (i.e., ${\cal K}$ = 3) similarity is selected as the image to be labeled next. (b) ${\cal K}$ number of labeled images with the highest similarities are selected. (c) Selected labeled images and corresponding segmentation maps are fused to obtain the pseudo segmentation map for the unlabeled image.} \label{fig2}
\end{figure}

The proposed pseudo-labeling method is built upon two hypotheses. First, inspired by curriculum learning~\cite{ref_curriculum}, we propose to label unlabeled images starting from the easy images to the difficult ones and we represent the easiness of an image as the similarity of that image to the labeled image pool. Second, we hypothesize that an unlabeled image can be segmented accurately using segmentation maps of the most similar labeled images of the unlabeled image. A schematic illustration of the proposed method is given in Fig.~\ref{fig2}.

The proposed method labels the unlabeled images in an increasing level of difficulty order, iteratively. At the beginning of each iteration, the method acquires the top ${\cal K}$\textsuperscript{th} similarity value between each unlabeled image and labeled image pool from the similarity matrix (Alg.~\ref{algo:pseudo}:3--7). The top ${\cal K}$\textsuperscript{th} similarity value for each unlabeled image is considered as the similarity between the unlabeled image and labeled image pool. The unlabeled image with the highest ${\cal K}$\textsuperscript{th} similarity value is selected as the image to be labeled next (Alg.~\ref{algo:pseudo}:8) (Fig.~\ref{fig2}(a)). Afterwards, the similarity between the selected unlabeled image and each labeled image is acquired from the same similarity matrix (Alg.~\ref{algo:pseudo}:9) and ${\cal K}$ number of labeled images with the highest similarities are selected to fuse their segmentation maps to produce a segmentation map for the unlabeled image (Alg.~\ref{algo:pseudo}:10) (Fig.~\ref{fig2}(b)). Before label fusion, selected labeled images and corresponding segmentation maps are aligned to the unlabeled image using the pre-computed inverse and forward transformations between images and the template to improve the accuracy of the label fusion. Finally, we use LOP STAPLE~\cite{ref_lopstaple} algorithm, which is a variant of the STAPLE~\cite{ref_staple} algorithm, to obtain label fusion of the most similar labeled images for the selected unlabeled image (Alg.~\ref{algo:pseudo}:11) (Fig.~\ref{fig2}(c)). LOP STAPLE algorithm computes voxelwise reliability weights for each segmentation map based on the intensity similarities in a pre-defined neighborhood between the target and corresponding template images. In our case, the unlabeled image is considered as the target image, and the most similar labeled images to the unlabeled image are considered as the template images. The voxelwise reliability weights of the segmentation maps associated with the template images are computed by measuring local similarities between the target and template images and the contribution of each segmentation map to label fusion is determined proportionally to its weights. The fused segmentation map is paired with the unlabeled image to add the pair to the labeled image pool and the pseudo-labeling process is being proceeded until no unlabeled image remains (Alg.~\ref{algo:pseudo}:12--15).

\begin{algorithm}[t]
\caption{{\sc Atlas-based Pseudo-Labeling}}\label{algo:pseudo}
\textbf{Input:} ${\cal N}$: set of expert labeled samples (set of $<$image, segmentation map$>$), ${\cal M}$: set of unlabeled samples ($<$image$>$), ${\cal S}$: similarity matrix, ${\cal K}$: similarity threshold \\
\textbf{Output:}  ${\cal N}$: set of expert and pseudo labeled samples (set of $<$image, segmentation map$>$)
\begin{algorithmic}[1]

\FOR{ $size({\cal M}$) $>$ 0 }
    \STATE ${\cal SIMS} = []$

\FOR{${m}$ $\in$ ${\cal M}$ }
    \STATE ${\cal SIM}_{m, \cal N } \gets$ ${\cal S}[m,\cal N]$
    \STATE ${\cal SIM}_{m, \cal K} \gets$ \sc SortAsc$({\cal SIM}_{m, \cal N})[\cal K]$ 
    \STATE ${\cal SIMS} \gets$ \sc Append$({\cal SIMS}, {\cal SIM}_{m, \cal K})$
\ENDFOR
\STATE ${\cal MSU} \gets$ \sc argmax$_1({\cal SIMS})$ 
\STATE ${\cal SIMS}_{{\cal MSU}, \cal N} \gets$ ${\cal S}[{\cal MSU},\cal N]$
\STATE $[{\cal MSS}_1, {\cal MSS}_2, ..., \cal MSS_K] \gets$ \sc argmax$_{1,...,\cal K}({\cal SIMS}_{{\cal MSU}, \cal N})$
\STATE ${\cal FUSION}_{\cal MSU} \gets$ \sc LopStaple$({\cal MSS}_1, {\cal MSS}_2, ..., \cal MSS_K)$ 
\STATE ${\cal N} \gets$ ${\cal N} \cup {\cal FUSION}_{\cal MSU}$ 
\STATE ${\cal M} \gets$ ${\cal M} \setminus {\cal MSU}$ 
\ENDFOR
\RETURN ${\cal N}$
\end{algorithmic}
\end{algorithm}
\setlength{\textfloatsep}{0.5cm}
\subsection{Deep Convolutional Neural Network Architecture and Training}
\label{sec:dcnn}

We propose to train a DCNN on the combination of expert and pseudo labeled images for each anatomical structure (class label), individually. Each proposed DCNN inputs a full-scale and full-resolution 3D brain MRI volume and outputs a binary segmentation map for a particular anatomical structure. The method employs a modified UNet~\cite{ref_unet} architecture with an encoder and decoder each with four consecutive spatial blocks connected by symmetric skip connections. Each convolution layer in the encoder uses $3 \times 3 \times 3$ kernels followed by a ReLU activation function and the output feature map is downsampled by a $2 \times 2 \times 2$ a max-pooling layer. The first convolutional block has 16 filters and the number of filters is doubled in each consecutive block. Correspondingly, the decoder employs similar convolution layers followed by a $2 \times 2 \times 2$ a upsampling layer. The architecture is finalized with a Sigmoid activation function for binary segmentation of each anatomical structure. The multi-class label for each voxel, $v$, is obtained by combining  ${\cal L}$ binary segmentation labels as follows:

\begin{equation}
\label{eqn:combine}
    {\cal S}_v=
    \begin{cases}
      0, & \text{if}\ max_1({\cal S}_{v,1}, {\cal S}_{v,2}, ..., {\cal S}_{v,{\cal L}}) < 0.5 \\
      argmax_1({\cal S}_{v,1}, {\cal S}_{v,2}, ..., {\cal S}_{v,{\cal L}}), & \text{otherwise}
    \end{cases}
  \end{equation}

\section{Results}
\label{sec:results}

We test the CORPS method on the Pediatric Imaging, Neurocognition, and Genetics (PING) data repository~\cite{ref_ping} that contains 750 T1-weighted 3D brain MRI volumes in a resolution of $256 \times 256 \times 160$ and corresponding expert segmentation maps consisting of 29 anatomical structures. We randomly divide the data collection into training and test sets including 600 and 150 volumes, respectively. We compare our method with the state-of-the-art UNet method~\cite{ref_unet} that is trained on the entire training set or a subset of the training set to measure the performance of the proposed approach. Note that we use the same architecture with the compared UNet method as our DCNN method for a fair comparison. 

\begin{figure}[t]
\includegraphics[width=\textwidth]{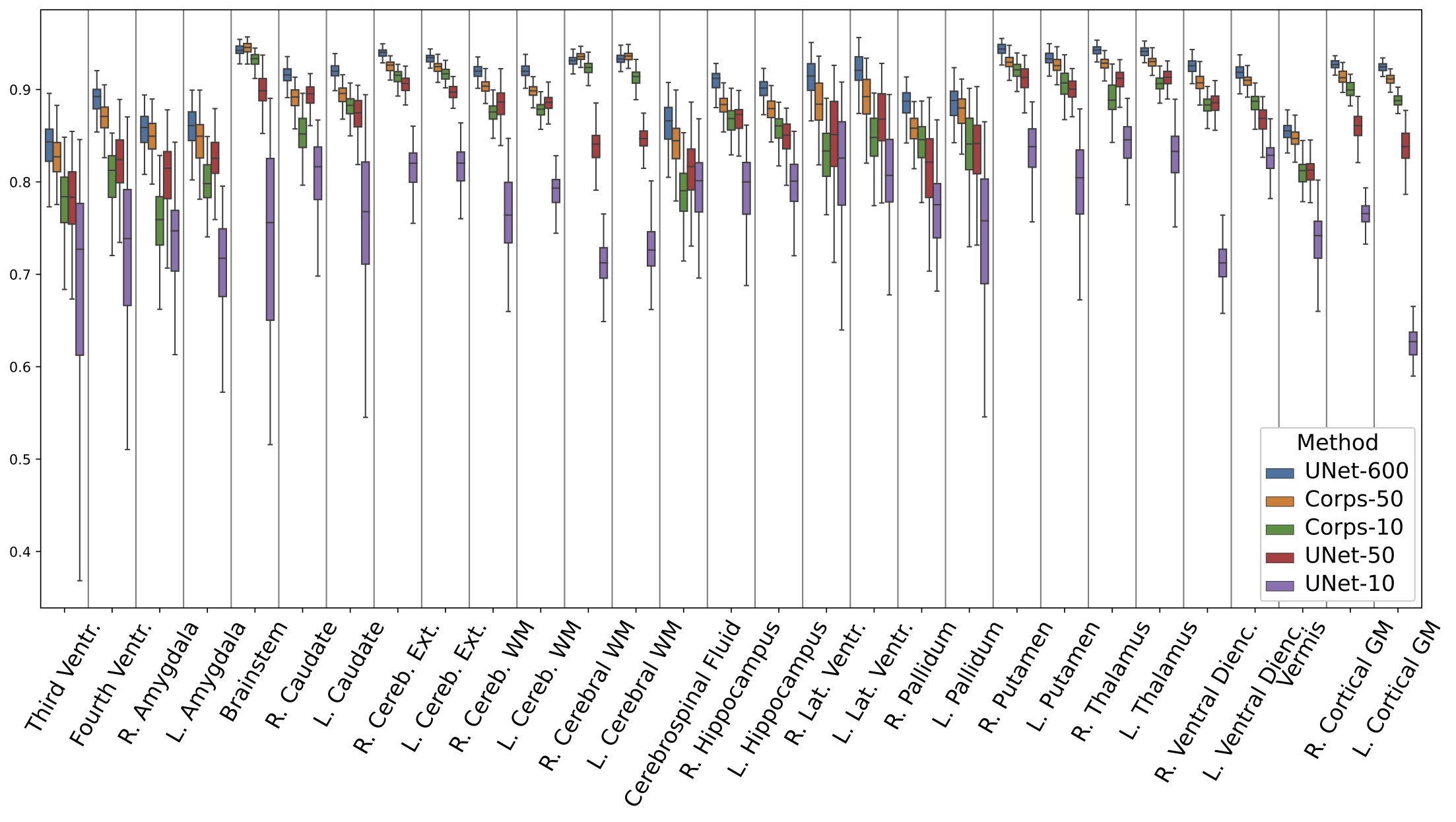}
\caption{Dice scores of CORPS and the comparison methods for all 29 structures.} \label{fig3}
\end{figure}

We generated cost-free pseudo-labels for each structure using the proposed method (CORPS) given $\sim$\%8 (50 expert labeled samples--\textit{CORPS-50}) and $\sim$\%2 (10 expert labeled samples--\textit{CORPS-10}) of the training set as expert labeled samples. Then, the DCNN modules of \textit{CORPS-50} and \textit{CORPS-10} are trained on their respective combination of expert and pseudo labeled samples. For comparisons, we also trained a UNet for each structure on the entire training set and consider this method as the upper bound performance of the comparison methods since it exploits 100\% of costly expert labeled data (\textit{UNet-600}). Lastly, we train a UNet for each structure on only the same $\sim$\%8 and $\sim$\%2 subsets of the expert labeled training set separately (\textit{UNet-50} and \textit{UNet-10}) to evaluate how much the proposed pseudo-labeling method improved the performance.

\begin{figure}[t]
\centering
\scriptsize{
\setlength{\tabcolsep}{0.1pt} 
\renewcommand{\arraystretch}{1} 
\begin{tabular}{ c c c c c c c }
\includegraphics[clip, trim=0.5cm 0.5cm 0.4cm 1.0cm, width =1.7cm]{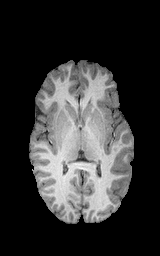} &
\includegraphics[clip, trim=0.5cm 0.5cm 0.4cm 1.0cm, width =1.7cm]{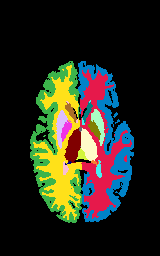} &
\includegraphics[clip, trim=0.5cm 0.5cm 0.4cm 1.0cm, width =1.7cm]{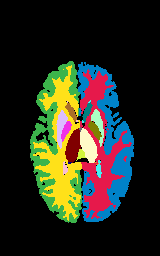} &
\includegraphics[clip, trim=0.5cm 0.5cm 0.4cm 1.0cm, width =1.7cm]{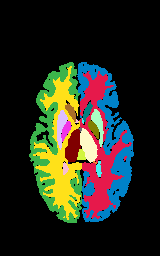} &
\includegraphics[clip, trim=0.5cm 0.5cm 0.4cm 1.0cm, width =1.7cm]{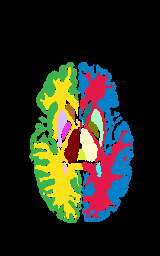} &
\includegraphics[clip, trim=0.5cm 0.5cm 0.4cm 1.0cm, width =1.7cm]{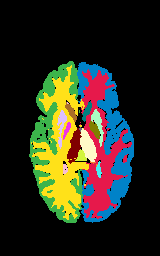} &
\includegraphics[clip, trim=0.5cm 0.5cm 0.4cm 1.0cm, width =1.7cm]{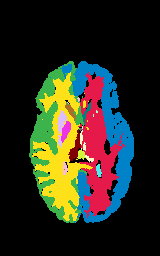} \\
Image &
Expert &
\textit{UNet-600} &
\textit{CORPS-50} &
\textit{CORPS-10} &
\textit{UNet-50} &
\textit{UNet-10}
\end{tabular}}
\caption{Qualitative results of CORPS and the comparison methods on a test volume.}
\label{fig4}
\end{figure}

Fig.~\ref{fig3} reports the box-plots of dice scores for the CORPS and its comparison methods. The results show that the \textit{CORPS-50} and \textit{CORPS-10} perform significantly better than \textit{UNet-50} and \textit{UNet-10}, respectively. This reveals the efficacy of incorporating unlabeled samples into the supervised training using the proposed atlas-based method. Yet more, \textit{CORPS-10} performs better than \textit{UNet-50} for almost half of the structures. This shows that a large number of accurate pseudo-labeled data can be superior to a small number of expert labeled data. Finally, the dice scores of \textit{CORPS-50}, which are very close to \textit{UNet-600}, reveal that the generated pseudo-labels are accurate and expert-level. Besides, the slight superiority of \textit{CORPS-50} over \textit{UNet-600} in the brainstem, left and right cerebral white matter structures can be interpreted as the proposed atlas-based pseudo-labeling can eliminate the variations in expert labeling procedures.

Fig.~\ref{fig4} shows the visual comparison of the CORPS with its counterparts. \textit{CORPS-50} and \textit{CORPS-10} predict segmentation maps marginally much similar to the expert map in comparison to \textit{UNet-50} and \textit{UNet-10}. While \textit{UNet-50} and \textit{UNet-10} suffer from having erosions in the subcortical structures or missegmenting some of them completely, and aggregating neighboring gyrus in the cortical gray matter, \textit{CORPS-50} and \textit{CORPS-10} accurately delineate these structures.

\vspace{-0.1cm}
\section{Conclusion}
\label{sec:conclusion}

We propose a semi-supervised segmentation framework that generates pseudo-labels in a cost-free atlas-based approach to improve the performance of its 3D supervised DCNN module. The results demonstrate the contribution of incorporating expert-level pseudo-labeled samples into supervised learning and employing lightweight and efficient full-scale and full-resolution 3D DCNNs to obtain a more context-aware model.

\end{document}